\documentclass[10pt,twocolumn,twoside]{IEEEtran}
\usepackage[utf8]{inputenc}         
\usepackage[english]{babel}
\usepackage{multirow}

\usepackage{multicol}

\usepackage{soul}

\usepackage[normalem]{ulem}

\usepackage{cite}

\ifCLASSINFOpdf
\usepackage[pdftex]{graphicx}
\usepackage{epstopdf}
\graphicspath{{./image/}}
\else
\fi
\usepackage[cmex10]{amsmath}
\usepackage{amssymb}
\usepackage{array}
\usepackage{rotating}

\DeclareMathOperator*{\argmax}{argmax}

\DeclareMathOperator*{\sigm}{sigm}

\usepackage{makecell}
\usepackage{subcaption}
\usepackage{textcomp}
\usepackage{url}

\usepackage{lipsum}

\usepackage{url}

\usepackage{multirow}
\usepackage{balance}
\usepackage{cite}
\usepackage{capt-of}
\usepackage[cmex10]{amsmath}
\usepackage{amssymb}

\usepackage{tabularx}
\usepackage{textcomp}
\usepackage{bm}
\usepackage{calc}

\usepackage{xcolor,colortbl}

\usepackage{tabu}
\usepackage{mdframed}

\makeatletter
\def\blfootnote{\xdef\@thefnmark{}\@footnotetext}
\makeatother

\usepackage[ruled,lined]{algorithm2e}

\begin{document}
	\bstctlcite{IEEEexample:BSTcontrol}
	\title{SeqSleepNet: End-to-End Hierarchical Recurrent Neural Network for Sequence-to-Sequence Automatic Sleep Staging}
	%
	%
	%

	\author{Huy~Phan$^*$,
		Fernando~Andreotti,
		Navin~Cooray,
		Oliver~Y.~Ch\'{e}n,
		and~Maarten~De~Vos
		\thanks{H. Phan is with the School of Computing, University of Kent, Chatham Maritime, Kent ME4 4AG, United Kingdom and the Institute of Biomedical Engineering, University of Oxford, Oxford OX3 7DQ, United Kingdom. F. Andreotti, N. Cooray, O. Y. Ch\'{e}n, and M. De Vos are with the Institute of Biomedical Engineering, University of Oxford, Oxford OX3 7DQ, United Kingdom.}
		\thanks{$^*$Corresponding author: {\tt\footnotesize h.phan@kent.ac.uk}}}
	
	%
	%

	\markboth{This Article Has Been Published in IEEE Transactions on Neural Systems and Rehabilitation Engineering}%
	{This Article Has Been Published in IEEE Transactions on Neural Systems and Rehabilitation Engineering}
	%



	\maketitle
	
	\begin{abstract}
		
		Automatic sleep staging has been often treated as a simple classification problem that aims at determining the label of individual target polysomnography (PSG) epochs one at a time. In this work, we tackle the task as a sequence-to-sequence classification problem that receives a sequence of multiple epochs as input and classifies all of their labels at once. For this purpose, we propose a hierarchical recurrent neural network named SeqSleepNet\footnote{Source code is available at \url{http://github.com/pquochuy/SeqSleepNet}}. At the epoch processing level, the network consists of a filterbank layer tailored to learn frequency-domain filters for preprocessing and an attention-based recurrent layer designed for short-term sequential modelling. At the sequence processing level, a recurrent layer placed on top of the learned epoch-wise features for long-term modelling of sequential epochs. The classification is then carried out on the output vectors at every time step of the top recurrent layer to produce the sequence of output labels. 
		Despite being hierarchical, we present a strategy to train the network in an end-to-end fashion. We show that the proposed network outperforms state-of-the-art approaches, achieving an overall accuracy, macro F1-score, and Cohen's kappa of $87.1\%$, $83.3\%$, and $0.815$ on a publicly available dataset with 200 subjects.
	\end{abstract}
	

	%
	\IEEEpeerreviewmaketitle
	
	\section{Introduction}
	\label{sec:intro}
	\blfootnote{\footnotesize DOI: 10.1109/TNSRE.2019.2896659}
	Humans spend around one-third of their lives sleeping, this process is crucial to protect the mental and physical health of an individual \cite{Siegel2005}. Sleep disorders are becoming an alarmingly common health problem, affecting millions of people worldwide. A survey conducted in the US between 1999 and 2004 reveals that 50-70 million adults suffer from over 70 different sleep disorders and 60 percent of adults report having sleep problems a few nights a week or more \cite{InstMed2006,Krieger2017}.
	
	Sleep scoring \cite{Iber2007, Hobson1969} is a fundamental step in sleep assessment and diagnosis and requires the analysis of 30-second polysomnography (PSG) epochs to determine their sleep stages. In clinical environments, sleep staging is mainly performed manually by human experts following developed guidelines \cite{Iber2007, Hobson1969}. The scoring procedure is labor-intensive, time-consuming, costly, and prone to human errors. Therefore, a large body of work aims to automate this task \cite{Redmond2006, Alickovic2018, Phan2018e, Supratak2017, Tsinalis2016, Mikkelsen2018, Stephansen2018, Chambon2018, Andreotti2018, Andreotti2018b}. Furthermore, there is an growing need of home-based sleep monitoring  \cite{Mikkelsen2018b,Looney2016,Goverdovsky2016,Kidmose2013} to provide scalable monitoring solutions that would benefit a greater population and provide a platform for epidemiological studies. In order to achieve this two primary ingredients are needed. First, user-friendly, comfortable, long-term capable, clinical-grade wearable Electroencephalography (EEG) devices are required. A number of such devices were developed and validated, such as in-ear EEG \cite{Kidmose2013, Looney2012, Goverdovsky2016} and around-the-ear EEG \cite{Mikkelsen2015,Mikkelsen2018b}. Second, reliable automatic sleep staging methods are equally indispensable.
	
	\begin{figure*} [!t]
		\centering
		\includegraphics[width=0.85\linewidth]{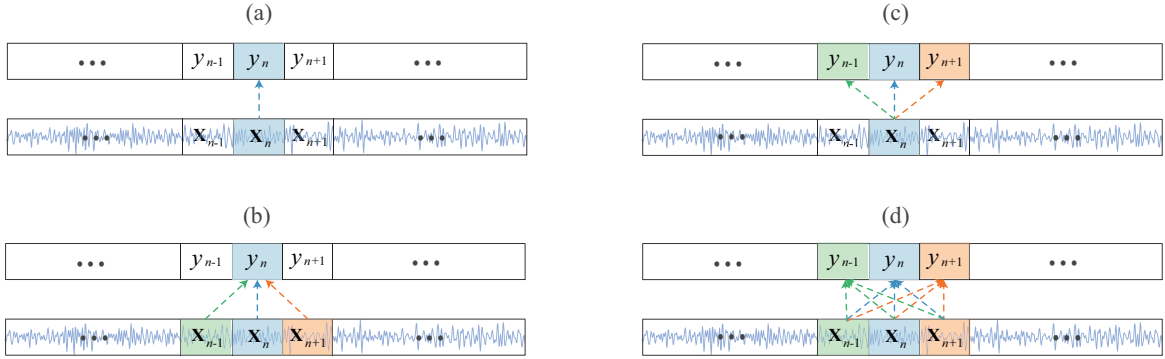}
		\caption{Illustration of the classification schemes used for automatic sleep staging. (a) one-to-one, (b) many-to-one, (c) one-to-many, and (d) the proposed many-to-many.}
		\label{fig:seq_to_seq_sleep_staging}
	\end{figure*}
	
	In the last few years, the research community has witnessed an influx of deep learning methods used for automatic sleep staging in replacement of conventional feature-based machine learning approaches. Deep learning methods offer several advantages over the conventional ones and have been successful in numerous other domains. First, since public sleep data are rapidly growing (i.e. hundreds to thousands of subjects are becoming a norm), deep networks are efficient in handling a large amount of data by repeatedly learning from small batches of data to converge to the final model. Second, their power in learning features automatically from low-level signals makes hand-crafting several intricate features no longer necessary. Several types of deep network architectures exist and have been proposed for automatic sleep scoring: Convolutional Neural Networks (CNNs) \cite{Phan2018e, Tsinalis2016, Mikkelsen2018, Chambon2018, Andreotti2018, Andreotti2018b}, Deep Belief Networks (DBNs) \cite{Laengkvist2012}, Auto-encoder \cite{Tsinalis2016b}, Deep Neural Networks (DNNs), and Recurrent Neural Networks (RNNs) \cite{phan2018d}. Combinations of different architectures, such as DNN+RNN \cite{Dong2017} and CNN+RNN \cite{Supratak2017,Stephansen2018} have also been exploited. With the deep learning methods evolving, automatic sleep staging performance has been boosted considerably as state-of-the-art results have been reported on several datasets \cite{Phan2018e, Stephansen2018, Chambon2018, Supratak2017}. 
	
	There are many ways to characterize existing works in automatic sleep staging, such as single-channel versus multi-channel and shallow learning vs deep learning. Here, we pursuit an approach that categorizes them into classification schemes based on the number of input epochs and output labels during classification. To this end, prior works can be grouped into \emph{one-to-one}, \emph{many-to-one}, \emph{one-to-many} schemes as illustrated in Figure \ref{fig:seq_to_seq_sleep_staging} (a)-(c), respectively. Following the one-to-one scheme, a classification model receives a single PSG epoch as input at a time and produces a single corresponding output label \cite{phan2018c, phan2018d, Andreotti2018, Andreotti2018b}. Although being straightforward, this classification scheme cannot take into account the existing dependency between PSG epochs \cite{Phan2018e,Iber2007,Sousa2015,Liang2011}. As an extension of the one-to-one, the many-to-one scheme augments the classification of a target epoch by additionally combining it with its surrounding epochs to make a \emph{contextual input}. This scheme has been the most widely used in prior works, not only those relying on more conventional methods \cite{Aboalayon2016,Patanaik2018} but also modern deep neural networks \cite{Mikkelsen2018, Supratak2017, Dong2017, Tsinalis2016, Tsinalis2016b, Chambon2018}. The work in \cite{Phan2018e} showed that while the \emph{contextual input} does not always lead to performance improvement regardless of the choice of classification model, it also suffers from the modelling ambiguity and high computational overhead. The one-to-many scheme is orthogonal to the many-to-one scheme and was recently proposed in \cite{Phan2018e} with the concept of \emph{contextual output}. Under this scheme, a multitask model receives a single target epoch as input and jointly determines both the target label and the labels of its neighboring epochs in the contextual output. This scheme is still able to leverage the inter-epoch dependency while avoiding the limitations of the contextual input in the many-to-one-scheme. More importantly, the underlying multitask model has the capability to produce an ensemble of decisions on a certain epoch which can be then aggregated to yield a more reliable final decision \cite{Phan2018e}. However, a common drawback of both many-to-one and one-to-many schemes is that they cannot accommodate a long context, e.g. tens of epochs. 
	
	In this work, we seek to overcome this major limitation and unify all aforementioned classification schemes with the proposed many-to-many approach illustrated in Figure \ref{fig:seq_to_seq_sleep_staging}(d). Our goal is to map an input sequence of multiple epochs to the sequence of all target labels at once. Therefore, the automatic sleep staging task is framed as a \emph{sequence-to-sequence} classification problem. With this generalized scheme, we can circumvent disadvantages of other schemes (i.e. short context, modelling ambiguity, and computational overhead) while maintaining the one-to-many's advantage regarding the availability of decision ensemble. It should be stressed that the sequence-to-sequence problem formulated here does not simply imply a set of one-to-one mappings between one epoch in the input sequence and its corresponding label in the output sequence. In contrast, due to the inter-epoch dependency, a label in the output sequence may inherently interact with all epochs in the input sequence via some intricate relationship that need to be modelled. To accomplish sequence-to-sequence classification we present \emph{SeqSleepNet}, a hierarchical recurrent neural network architecture. SeqSleepNet is composed of three main components: (1) parallel \emph{filterbank layers} for preprocessing, (2) an epoch-level bidirectional RNN coupled with the attention mechanism for short-term (i.e. intra-epoch) sequential modelling, and (3) a sequence-level bidirectional RNN for long-term (i.e. inter-epoch) sequential modelling. The network is trained in an end-to-end manner. End-to-end network training is desirable in deep learning as an end-to-end network learns the global solution directly in contrast to multiple-stage training that estimates local solutions in separate stages. The power of end-to-end learning has been proven many times in various domains \cite{Bojarski2016,Collobert2011,Silver2016,Krizhevsky2012, Mnih2015, Levine2016}. Moreover, end-to-end training is more convenient and elegant.
	
	Our proposed method bears resemblance to some existing works. Learning data-driven filters with a filterbank layer has been shown to be efficient in our previous works \cite{phan2018c,phan2018d,Phan2018e}. However, instead of training a filterbank layer separately with a DNN, here multiple filterbank layers for multichannel input are parts of the classification network and are trained end-to-end. There also exists a few multiple-output network architectures proposed for automatic sleep staging, nevertheless, they are either limited to accommodate a long-term context \cite{Phan2018e} or need to be trained in multiple stages rather than end-to-end \cite{Supratak2017,Dong2017}. In addition, these works used CNNs or DNNs for epoch-wise feature learning while, in the proposed SeqSleepNet, we employ a recurrent layer coupled with the attention mechanism for this purpose. Given the sequential nature of sleep data, the sequential modelling capability of RNNs \cite{Hochreiter1997,Cho2014} make them potential candidates for this purpose but have been left uncharted. On one hand, we demonstrate that the sequential features learned with the attention-based recurrent layer result in a better performance than the convolutional ones. On the other hand, using our end-to-end training strategy, we also build end-to-end variants of these multiple-output networks as baselines and show that the proposed DeepSleepNet significantly outperforms all these baselines and set state-of-the-art performance on the experimental dataset.
	
	\section{Montreal Archive of Sleep Studies (MASS) Dataset}
	\label{sec:datasets}
	
	The public dataset Montreal Archive of Sleep Studies (MASS) \cite{Oreilly2014} was used for evaluation. MASS is a considerably large open-source dataset which were pooled from different hospital-based sleep laboratories. It consists of whole-night recordings from 200 subjects aged between 18-76 years (97 males and 103 females), divided into five subsets (SS1 - SS5). Each epoch of the recordings was manually labelled by experts  according to the AASM standard \cite{Iber2007} (SS1 and SS3) or the R\&K standard \cite{Hobson1969}  (SS2, SS4, and SS5). We converted different annotations into five sleep stages \{W, N1, N2, N3, and REM\} as suggested in \cite{Imtiaz2014,Imtiaz2015}. Furthermore, those recordings with 20-second epochs were converted into 30-second ones by including 5-second segments before and after each epoch. In our analysis, we used the entire dataset (i.e. all five subsets), following the experimental setup suggested in \cite{Phan2018e}. Apart from an EEG channel, an EOG and EMG channel were included to complement the EEG as they have been shown to be valuable addition sources for automatic sleep staging \cite{Phan2018e, Chambon2018, Lajnef2015, Huang2014, Mikkelsen2018, Andreotti2018, Stephansen2018}. We adopted and studied combinations of the C4-A1 EEG, an average EOG (ROC-LOC), and an average EMG (CHIN1-CHIN2) channels in our experiments. The signals, originally sampled at 256 Hz, were downsampled to 100 Hz.
	
	\section{SeqSleepNet: End-to-End Hierarchical Recurrent Neural Network}
	\label{sec:framework}
	
\begin{figure*} [!t]
		\centering
		\includegraphics[width=0.9\linewidth]{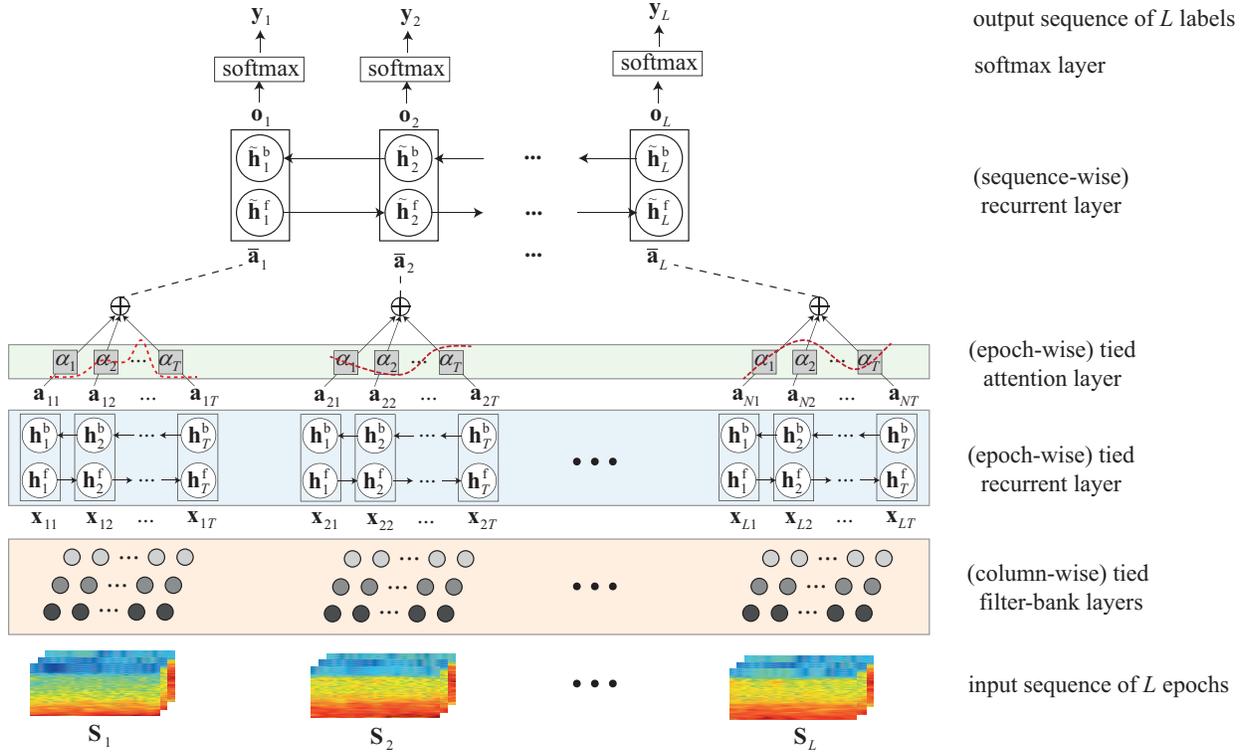}
		\caption{Illustration of SeqSleepNet, the proposed end-to-end hierarchical RNN for sequence-to-sequence sleep staging.}
		\label{fig:arnn}
	\end{figure*}

	The proposed SeqSleepNet for sequence-to-sequence sleep staging is illustrated in Figure \ref{fig:arnn}. Formally, given a sequence of PSG epochs of length $L$ represented by $(\mathbf{S}_1, \mathbf{S}_2, \ldots, \mathbf{S}_L)$, the goal is to compute  a sequence of outputs $(\mathbf{y}_1, \mathbf{y}_2, \ldots, \mathbf{y}_L)$ that maximizes the conditional probability $p(\mathbf{S}_1, \mathbf{S}_2, \ldots, \mathbf{S}_L\,|\,\mathbf{y}_1, \mathbf{y}_2, \ldots, \mathbf{y}_L)$.
	
	An epoch in the input sequence consisting of $C$ channels (i.e. EEG, EOG, and EMG in this work), are firstly transformed into a time-frequency image $\mathbf{S}$ of $C$ image channels. Parallel filterbank layers \cite{phan2018c, phan2018d} are tailored to learn channel-specific frequency-domain filterbanks to preprocess the input image for frequency smoothing and dimension reduction. Furthermore, after channel-specific preprocessing, all image channels are concatenated in the frequency direction to form an image $\mathbf{X}$. The image $\mathbf{X}$ itself can be interpreted as a sequence of feature vectors which correspond to the image columns. The epoch-level attention-based bidirectional RNN is then used to encode the feature vector sequence of the epoch into a fixed attentional feature vector $\mathbf{\bar{a}}$. Finally, the sequence of attentional feature vectors $\mathbf{\bar{A}}=(\mathbf{\bar{a}}_1, \mathbf{\bar{a}}_2, \ldots, \mathbf{\bar{a}}_L)$ obtained from the input epoch sequence are modelled by the sequence-level bidirectional RNN situating on top of the network hierarchy to compute the output sequence $\mathbf{\hat{Y}}=(\mathbf{\hat{y}}_1, \mathbf{\hat{y}}_2, \ldots, \mathbf{\hat{y}}_L)$.
	
	It should be noted that, in the SeqSleepNet, the filterbank layers are tied (i.e. shared parameters) between all epochs' local features (i.e. spectral image columns) and the epoch-level attention-based bidirectional RNN layer are tied between all epochs in the input sequence. 
	
	\subsection{Time-Frequency Image Representation}
	\label{ssec:representation}
	
	The constituent signals of a 30-second PSG epoch (i.e. EEG, EOG, and EMG) are transformed into power spectra via short-time Fourier transform (STFT) with a window size of two seconds and 50\% overlap. Hamming window and 256-point Fast Fourier Transform (FFT) are used. Logarithm scaling is then applied to the spectra to convert them into log-power spectra. As a result, a multi-channel image $\mathbf{S} \in \mathbb{R}^{F \times T \times C}$ is obtained where $F=129$, $T=29$, and $C=3$ denote the number of frequency bins, the number of spectral columns (i.e. time indices), and the number of channels.
	
	\subsection{Filterbank Layers}
	\label{ssec:filterbank}
	
	We tailor a filterbank layer for learning frequency-domain filterbanks as in our previous works \cite{phan2018c, phan2018d}. The learned filterbank is expected to emphasize the subbands that are more important for the task at hand and  attenuate those less important. However, instead of training a separate DNN for this purpose, the filterbank layers are parts of the classification network SeqSleepNet and are learned end-to-end. Moreover, due to the different signal characteristics of EEG, EOG, and EMG, it is reasonable to learn $C$ channel-specific filterbanks with $C$ separate filterbank layers.
	
	Considering the $c$-th filterbank layer with respect to the $c$-th image channel $\mathbf{S}^c \in \mathbb{R}^{F \times T}$ where $1 \le c \le C$ and assuming that we want to learn a frequency-domain filertbank of $M$ filters where $M < F$, the filterbank layer in principle is a fully-connected layer of $M$ hidden units. The weight matrix $\mathbf{W}^c \in \mathbb{R}^{F \times M}$ of this layer plays the role of the filterbank's weight matrix. Since a filterbank has characteristics of being non-negative, band-limited, and ordered in frequency, it is necessary to enforce the following constraints \cite{Yu2017} for the learned filterbank to have these characteristics: 
	\begin{align}
	\mathbf{W}^c_{\text{fb}} = f_{+}(\mathbf{W}) \odot \mathbf{T}.
	\label{eq:wfb}
	\end{align}
	Here, $f_{+}$ denotes a non-negative function to make the elements of $\mathbf{W}$ non-negative, in this study the \emph{sigmoid} function is adopted. $\mathbf{T} \in \mathbb{R}^{F \times M}_{+}$ is the constant non-negative matrix to enforce the filters to have limited band, regulated shape and ordered by frequency. Similar to \cite{phan2018c}, we employ a linear-frequency triangular filterbank matrix for $\mathbf{T}$.	The $\odot$ operator denotes the element-wise multiplication. 
	
	Presenting the image $\mathbf{S}^c$ to the filterbank layer, we obtained an output image $\mathbf{X}^c \in \mathbb{R}^{M \times T}$ given by
	\begin{align}
	\mathbf{X}^c = {\mathbf{W}_{\text{fb}}^c}^\mathsf{T}\mathbf{S}^c.
	\label{eq:wfb2}
	\end{align}
	All together, filtering the $C$-channel input image $\mathbf{S}~\in~\mathbb{R}^{F \times T \times C}$ in frequency direction with $C$ filterbank layers results in the $C$-channel output image $\mathbf{X} \in \mathbb{R}^{M \times T \times C}$ which has smaller size in frequency dimension. Eventually, we concatenate the image channels of $\mathbf{X}$ in frequency direction to make $\mathbf{X}$ a single-channel image of size $MC \times T$.
	
	\subsection{Short-term Sequential Modelling}
	\label{ssec:shortterm_rnn}
	
	Many approaches to extract features that represent an epoch exist. Apart from a large body of hand-crafted features \cite{Aboalayon2016}, automatic feature learning with deep learning approaches are becoming more common \cite{Laengkvist2012, phan2018c, phan2018d, Mikkelsen2018, Stephansen2018, Tsinalis2016, Tsinalis2016b, Supratak2017, Dong2017, Andreotti2018, Koch2018a, Koch2018b, Phan2017}. Here, we employ a bidirectional RNN coupled with the attention mechanism \cite{Luong2015b,Bahdanau2015} to learn sequential features for epoch representation. Due to the RNN's sequential modelling capability, it is expected to capture temporal dynamics of input signals to produce good features \cite{phan2018d}.
	
	For convenience, we interpret the image $\mathbf{X}$ after the filterbank layers as a sequence of $T$ feature vectors $\mathbf{X} \equiv (\mathbf{x}_1, \mathbf{x}_2, \ldots, \mathbf{x}_T)$ where each $\mathbf{x}_t \in \mathbb{R}^{MC}$, $1 \le t \le T$, is the image column at time index $t$.
	We then aim to read the sequence of feature vectors into a single feature vector using the attention-based bidirectional RNN. 
	
	The forward and backward recurrent layers of the RNN iterate over individual feature vectors of the sequence in opposite directions and compute forward and backward sequences of hidden state vectors $\mathbf{H}^{\text{f}}~=~(\mathbf{h}^{\text{f}}_1, \mathbf{h}^{\text{f}}_2, \ldots, \mathbf{h}^{\text{f}}_T)$ and $\mathbf{H}^{\text{b}}=(\mathbf{h}^{\text{b}}_1, \mathbf{h}^{\text{b}}_2, \ldots, \mathbf{h}^{\text{b}}_T)$, respectively, where
	\begin{align}
	\mathbf{h}^{\text{f}}_t &= \mathcal{H}(\mathbf{x}_t\,, \mathbf{h}^{\text{f}}_{t-1}), \label{eq:rnn_hidden_forward} \\
	\mathbf{h}^{\text{b}}_t &= \mathcal{H}(\mathbf{x}_t\,, \mathbf{h}^{\text{b}}_{t+1}), \mbox{~} 1 \le t \le T.
	\label{eq:rnn_hidden_backward}
	\end{align}
	In (\ref{eq:rnn_hidden_forward}) and (\ref{eq:rnn_hidden_backward}),  $\mathcal{H}$ denotes the hidden layer function. Long Short-Term Memory (LSTM) \cite{Hochreiter1997} and Gated Recurrent Unit (GRU) cell \cite{Cho2014} are most commonly used for $\mathcal{H}$. LSTM cell and GRU cell have been shown to perform comparably on many machine learning tasks, however, the latter has less parameters and is therefore more computational-efficient than the former \cite{Chung2014}. Here, we employ the latter which is implemented by the following functions:
	\begin{align}
	\mathbf{r}_t &= \sigm\left( \mathbf{W}_{sr}\mathbf{s}_t + \mathbf{W}_{hr}\mathbf{h}_{t-1} + \mathbf{b}_r\right), \\
	\mathbf{z}_t &= \sigm\left( \mathbf{W}_{sz}\mathbf{s}_t + \mathbf{W}_{hz}\mathbf{h}_{t-1} + \mathbf{b}_z\right), \\
	\mathbf{\bar{h}}_t &= \tanh\left( \mathbf{W}_{sh}\mathbf{s}_t + \mathbf{W}_{hh}\left(\mathbf{r}_t \odot \mathbf{h}_{t-1}\right) + \mathbf{b}_h\right), \\
	\mathbf{h}_t &= \mathbf{z}_t \odot \mathbf{h}_{t-1} + (1 - \mathbf{z}_t) \odot \mathbf{\bar{h}}_t,
	\end{align}
	where the $\mathbf{W}$ variables denote the weight matrices and the $\mathbf{b}$ variables are the biases. The $\mathbf{r}$, $\mathbf{z}$, and $\mathbf{\bar{h}}$ variables represent the reset gate vector, the update gate vector, and the new hidden state vector candidate, respectively. 
	
	The RNN produces the sequence of output vectors $\mathbf{A}=(\mathbf{a}_1, \mathbf{a}_2, \ldots, \mathbf{a}_T)$ where $\mathbf{a}_t$ is computed as
	\begin{align}
	\mathbf{a}_t &= \mathbf{W}_{ha}[\mathbf{h}^{\text{b}}_t \oplus \mathbf{h}^{\text{f}}_t] + \mathbf{b}_{a},
	\label{eq:rnn_output}
	\end{align}
	where $\oplus$ represents vector concatenation.
	
	The attention layer \cite{Luong2015b,Bahdanau2015} is then used to learn a weighting vector to combine these output vectors at different time steps into a single feature vector. The rationale is that those parts of the sequence which are more informative should be associated with strong weights and vice versa. Formally, the attention weight $\alpha_t$ at the time index $t$ is computed as
	\begin{align}
	\alpha_t = \frac{\exp\left(f(\mathbf{a}_t)\right)}{\sum^{T}_{i=1}\exp\left(f(\mathbf{a}_i)\right)}.
	\label{eq:attention_weight}
	\end{align}
	Here, $f$ denotes the scoring function of the attention layer and is given by
	\begin{align}
	f(\mathbf{a}) = \mathbf{a}^\mathsf{T}\mathbf{W}_{\text{att}},
	\label{eq:attention_layer}
	\end{align}
	where $\mathbf{W}_{\text{att}}$ is the trainable weight matrix. The attentional feature vector $\mathbf{\bar{a}}$ is obtained as a weighting combination of the recurrent output vectors:
	\begin{align}
	\mathbf{\bar{a}} = \sum^{T}_{t=1}\alpha_i\mathbf{a}_t.
	\label{eq:attentiive_output}
	\end{align}
	
	The attentional feature vector $\mathbf{\bar{a}}$ is used as the representation of the PSG epoch in the next sequence-level modelling.
	
	\subsection{Long-term Sequential Modelling}
	\label{ssec:longterm_rnn}
	
	Processing the input sequence $(\mathbf{S}_1, \mathbf{S}_2, \ldots, \mathbf{S}_L)$ with the filterbank layers in Section \ref{ssec:filterbank} and the attention-based bidirectional RNN layer in Section \ref{ssec:shortterm_rnn}
	results in a sequence of attentional feature vectors $\mathbf{\bar{A}}=(\mathbf{\bar{a}}_1, \mathbf{\bar{a}}_2, \ldots, \mathbf{\bar{a}}_L)$ where ${\mathbf{\bar{a}}}_l$, $1 \le l \le L$, is given in (\ref{eq:attentiive_output}). The sequence-level bidirectional RNN is then used to model the sequence of epoch-wise feature vectors to encode long-term sequential information across epochs. Similar to the bidirectional RNN used for short-term sequential modelling in Section \ref{ssec:shortterm_rnn}, its forward and backward sequences of hidden state vectors $\mathbf{\tilde{H}}^{\text{f}}=(\mathbf{\tilde{h}}^{\text{f}}_1, \mathbf{\tilde{h}}^{\text{f}}_2, \ldots, \mathbf{\tilde{h}}^{\text{f}}_L)$ and $\mathbf{\tilde{H}}^{\text{b}}=(\mathbf{\tilde{h}}^{\text{b}}_1, \mathbf{\tilde{h}}^{\text{b}}_2, \ldots, \mathbf{\tilde{h}}^{\text{b}}_L)$ are computed using (\ref{eq:rnn_hidden_forward}) and (\ref{eq:rnn_hidden_backward}) with $\mathbf{\bar{A}}=(\mathbf{\bar{a}}_1, \mathbf{\bar{a}}_2, \ldots, \mathbf{\bar{a}}_L)$ now playing the role of the input sequence. Again, GRU cells \cite{Cho2014} are used for its forward and backward recurrent layers.
	
	The sequence of output vectors $\mathbf{O}~=~(\mathbf{o}_1, \mathbf{o}_2, \ldots, \mathbf{o}_L)$ is then obtained where $\mathbf{o}_l$, $1 \le l \le L$, is computed as 
	\begin{align}
	\mathbf{o}_l &= \mathbf{\tilde{W}}_{ho}[\mathbf{\tilde{h}}^{\text{b}}_l \oplus \mathbf{\tilde{h}}^{\text{f}}_l] + \mathbf{\tilde{b}}_{o}.
	\label{eq:rnn_output2}
	\end{align}
	Each output vector $\mathbf{o}_l$ is presented to a softmax layer for classification to produce the sequence of classification outputs $\mathbf{\hat{Y}}=(\mathbf{\hat{y}}_1, \mathbf{\hat{y}}_2, \ldots, \mathbf{\hat{y}}_L)$, where $\mathbf{\hat{y}}_l$ is a output probability distribution over all sleep stages.
	
	\subsection{Sequence Loss}
	In the proposed sequence-to-sequence setting, we want to penalize the network for misclassification of any element of an input sequence. Given the input sequence $(\mathbf{S}_1, \mathbf{S}_2, \ldots, \mathbf{S}_L)$ with the ground-truth one-hot encoding vectors $(\mathbf{y}_1, \mathbf{y}_2, \ldots, \mathbf{y}_L)$ and the corresponding sequence of classification outputs $(\mathbf{\hat{y}}_1, \mathbf{\hat{y}}_2, \ldots, \mathbf{\hat{y}}_L)$, the sequence loss reads as follows (note that the sequence loss $E^s$ is normalized by the sequence length $L$):
	\begin{align}
	E^s(\bm{\theta}) = -\frac{1}{L}\sum_{l=1}^{L} \mathbf{y}_l\log\left(\mathbf{\hat{y}}_l\left(\bm{\theta}\right)\right).
	\label{eq:sequence_loss}
	\end{align}
	
	The network is trained to minimize the sequence loss over $N$ training sequences in the training data:
	\begin{align}
	E(\bm{\theta}) = -\frac{1}{N}\sum_{n=1}^{N}E_n^s(\bm{\theta}) + \frac{\lambda}{2}\|\bm{\theta}\|^2_2,
	\label{eq:sequence_loss_regularized}
	\end{align}
	where $E_n^s$ is given in (\ref{eq:sequence_loss}). Here, $\lambda$ denotes the hyper-parameter that trades off the error terms and the $\ell_2$-norm regularization term. 
	
	\subsection{End-to-End Training Details}
	\label{ss:end_to_end_training}
	
	In the proposed SeqSleepNet, the input unit of a filterbank layer is a spectral column of an epoch's time-frequency image, that of the epoch-level attention-based bidirectional RNN is such an entire image, and that of the top sequence-level RNN is a sequence of attentional feature vectors encoding the input epoch sequence. In order to train the network end-to-end, we adaptively manipulate the input data, i.e. folding and unfolding, at different levels of the network hierarchy.
	
	For simplicity, let us assume the single-channel input, and therefore, the network has only one filterbank layer. Since the network, in practice, is trained with a mini batch of data at a time, assume that at each training iteration we use a mini-batch of $S$ sequences, each consists of $L$ epochs. For a recall, each epoch itself is represented by an time-frequency image of size $T \times F$ (cf. Section \ref{ssec:representation}) which will be interpreted as a sequence of $T$ image columns when necessary. We firstly unfold the $S$ input sequences to make a set of $S \times L \times T$ image columns, each of size $F$, to present to the filterbank layer. After the filterbank layer, we obtain a set of $S \times L \times T$ image columns but now each has a size of $M$. This set of image columns are then folded to form a set of $S \times L$ images, each of size $T \times M$, to feed into the epoch-level attention-based bidirectional RNN. This layer encodes each image into an attentional feature vector, resulting in a set of $S \times L$ such feature vectors. Eventually, this set of feature vectors are folded into a set of $S$ sequences, each consists of $L$ attentional feature vectors, to present to the sequence-level bidirectional RNN for sequence-to-sequence classification.
	
	\section{Ensemble of Decisions and Probabilistic Aggregation}
	\label{sec:aggregation}
	Since SeqSleepNet is a multiple-output network, advancing the input sequence of size $L$ by one epoch when evaluating it on a test recording will result in an ensemble of $L$ decisions at every epoch (except those at the recording's ends). Fusing this decision ensemble leads to a final decision which are usually better than individual ones \cite{Phan2018e}.
	
	We use the multiplicative aggregation scheme which are shown in \cite{Phan2018e} to be efficient for this purpose. The final posterior probability of a sleep stage $y_t \in \mathcal{L} = \{\text{W}, \text{N1}, \text{N2}, \text{N3}, \text{REM}\}$ at a time index $t$ is given by
	\begin{align}
	P(y_t) = \frac{1}{L} \prod_{i=t-L+1}^{t} P(y_t\,|\,\mathcal{S}_i). \label{eq:multiplicative_smoothing}
	\end{align}
	where $\mathcal{S}_i = (\mathbf{S}_i, \mathbf{S}_{i+1}, \ldots, \mathbf{S}_{L-1})$ is the epoch sequence starting at $i$. In order to avoid possible numerical problems when the ensemble size is large, it is necessary to carry out the aggregation in the logarithm domain. The equation (\ref{eq:multiplicative_smoothing}) is then re-written as
	\begin{align}
	\log P(y_t) = \frac{1}{L} \sum_{i=t-L+1}^{t} \log P(y_t\,|\,\mathcal{S}_i). \label{eq:multiplicative_smoothing_log}
	\end{align}
	
	Eventually, the predicted label $\hat{y}_t$ is determined by likelihood maximization:
	\begin{align}
	\hat{y}_t = \argmax_{y_t} \log P(y_t) \text{~for~} y_t \in \mathcal{L}. \label{eq:likelihood_maximization}
	\end{align}

	\section{Experiments}
	\label{sec:experiments}
	
	\subsection{Experimental Setup}
	We conducted 20-fold cross validation on the MASS dataset. At each iteration, 200 subjects were split into training, validation, and test set with 180, 10, and 10 subjects, respectively. 
	During training, we evaluated the network after every 100 training steps and the one yielded the best overall accuracy on the validation set was retained for evaluation.
	The outputs of 20 cross-validation folds were pooled and considered as a whole for computing the sleep staging performance.
	
	\subsection{Network Parameters}
	
	The network was implemented using \emph{TensorFlow  v1.3.0} framework \cite{Abadi2016}. The network parameters are shown in Table \ref{tab:network_parameters}. Particularly, we experimented with different sequence length of $\{10, 20, 30\}$ epochs, which is equivalent to $\{5, 10, 15\}$ minutes, to study its influence. The network was trained for 10 epochs with a minibatch size of 32 sequences. The sequences were sampled from the PSG recordings with a maximum overlapping (i.e. $L-1$ epochs), in this way, we generated all possible epoch sequences from the training data. 
	
	Beside $\ell_2$-norm regularization in (\ref{eq:sequence_loss_regularized}), dropout \cite{Srivastava2014} was employed for further regularization. Recurrent batch normalization \cite{Cooijmans2016} was also integrated to the GRU cell to improve its convergence. The network training was performed using \emph{Adam} optimizer \cite{Kingma2015} with a learning rate of $10^{-4}$. 
	
	\begin{table}[t!]
		\centering
		\caption{Parameters of the proposed network.}
		\begin{tabular}{|>{\arraybackslash}m{1.3in}|>{\centering\arraybackslash}m{0.65in}|}
			\hline
			{\bf Parameter} & {\bf Value} \parbox{1pt}{\rule{0pt}{1ex+\baselineskip}} \\ [0ex]  	
			\hline
			Sequence length $L$ & $\{10, 20, 30\}$ \parbox{1pt}{\rule{0pt}{0.5ex+\baselineskip}} \\ [0ex]
			Number of filters $M$ & 32 \parbox{1pt}{\rule{0pt}{0.5ex+\baselineskip}} \\ [0ex]
			Size of hidden state vector & 64 \parbox{1pt}{\rule{0pt}{0.5ex+\baselineskip}} \\ [0ex]
			Size of the attention weights & 64 \parbox{1pt}{\rule{0pt}{0.5ex+\baselineskip}} \\ [0ex]
			Dropout rate & $0.25$ \parbox{1pt}{\rule{0pt}{0.5ex+\baselineskip}} \\ [0ex]
			Regularization parameter $\lambda$ & $10^{-3}$ \parbox{1pt}{\rule{0pt}{0.5ex+\baselineskip}} \\ [0ex]
			\hline
		\end{tabular}
		\label{tab:network_parameters}
	\end{table}
	
	\subsection{Baseline Networks}
	\begin{figure*} [!t]
		\centering
		\begin{subfigure}[b]{.25\textwidth}
			\vfill
			\centering
			\includegraphics[width=1\linewidth]{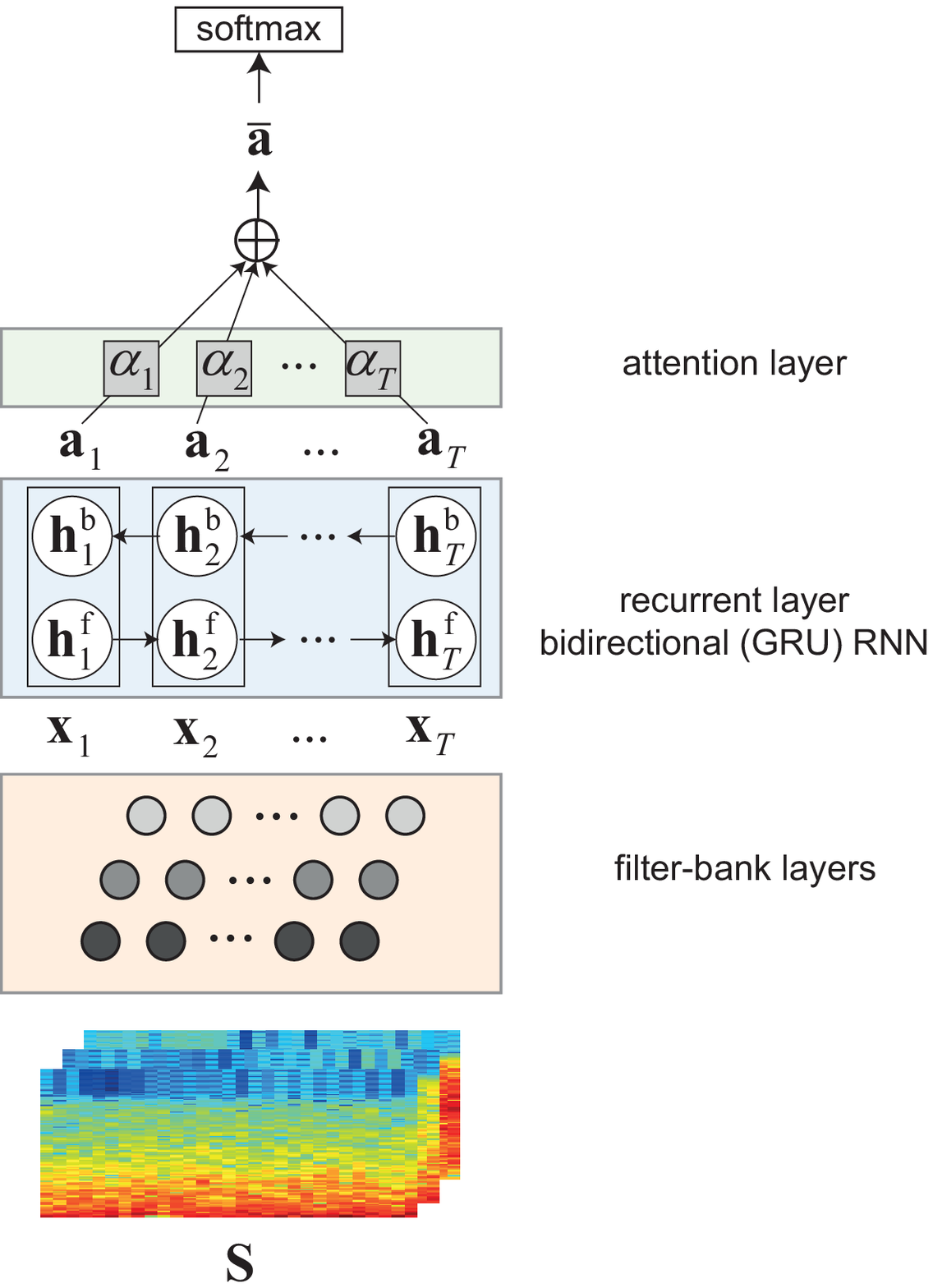}
			\captionof{figure}{End-to-end ARNN baseline.}
			\label{fig:ARNN}
		\end{subfigure}%
		\quad
		\quad
		\quad	
		\centering
		\begin{subfigure}[b]{0.65\textwidth}
			\includegraphics[width=1\linewidth]{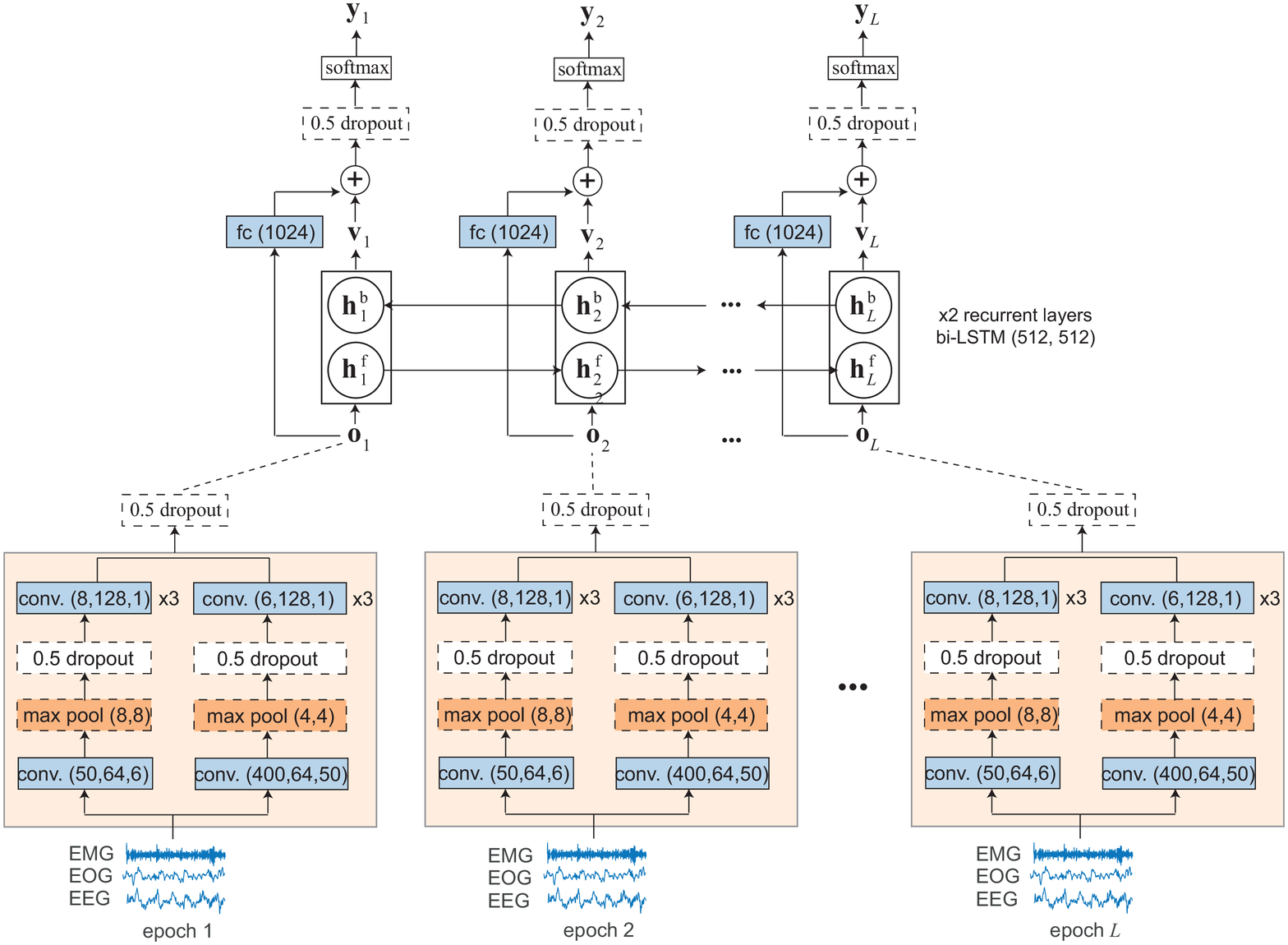}
			\captionof{figure}{End-to-end DeepSleepNet baseline.}
			\label{fig:deepsleepnet_endtoend}
		\end{subfigure}%
		\caption{Illustration of the developed baselines. In (b), \emph{conv. (n,w,s)} denotes a convolutional layer with \emph{n} 1-D filters of size \emph{w} and stride \emph{s}. \emph{max pool. (w,s)} denotes a 1-D max pooling layer with kernel size \emph{w} and stride \emph{s}. \emph{fc (n)} represents a fully connected layer with \emph{n} hidden units. Finally, \emph{bi-LSTM (n,m)} represents a bidirectional LSTM cell with size of its forward and backward hidden state vectors of \emph{n} and \emph{m}, respectively. Further details of these parameters can be found in \cite{Supratak2017}.}
	\end{figure*}
	
	In order to assess the efficiency of the proposed SeqSleepNet, apart from existing works, we developed three novel end-to-end baseline networks\footnote{Source code is available at \url{http://github.com/pquochuy/SeqSleepNet}} for comparison:
	
	{\bf End-to-end ARNN (E2E-ARNN)}: As illustrated in Figure \ref{fig:ARNN}, E2E-ARNN is the combination of the filterbank layers and the epoch-level attention-based bidirectional RNN of the proposed SeqSleepNet, and therefore,  is purposed for short-term sequential modelling. The objective is to assess the efficacy of the attention-based bidirectional RNN in epoch-wise feature learning. This baseline follows the standard one-to-one classification scheme, receiving a single epoch as input and outputting the corresponding sleep stage. The classification is accomplished by presenting the attentional output to a softmax layer. The network was trained with the standard cross-entropy loss. A similar attention-based bidirectional RNN was demonstrated to achieve good performance on a single-channel EEG setting in our previous work \cite{phan2018c}. However, here the filterbank learning and the sleep stage classification are jointly learned in an end-to-end manner. We used similar parameters as the SeqSleepNet's epoch-level processing block, except for the size of the attention weights which was set to 32. In addition, the network was trained for 20 epochs and was validated every 500 steps during training. 
	
	{\bf Multitask E2E-ARNN}: Inspired by multitask networks for sleep staging in \cite{Phan2018e}, this multitask network extends the E2E-ARNN baseline above to jointly determine the label of the input epoch and to predict the labels of its neighboring epochs. Therefore, this multiple-output baseline offers ensemble of decisions which was aggregated using the method described in Section \ref{sec:aggregation}. We used a context output size of 3 as in \cite{Phan2018e}.
	
	{\bf End-to-end DeepSleepNet (E2E-DeepSleepNet)}: Supratak \emph{et al.} \cite{Supratak2017} recently proposed DeepSleepNet and reported good performance on the MASS's subset SS3 with 62 subjects. This network comprises a deep CNN for epoch-wise feature learning topped up with a deep bidirectional RNN for capturing stage transitions. As described in \cite{Supratak2017}, these two parts were trained in two separate stages to yield good performance. Here, we developed an \emph{end-to-end} variant of DeepSleepNet, illustrated in Figure \ref{fig:deepsleepnet_endtoend}, and trained the model end-to-end using a similar strategy described in Section \ref{ss:end_to_end_training}. We will show that E2E-DeepSleepNet achieves a comparable performance (if not better) as that reported in \cite{Supratak2017}. The network parameters were kept as in the original version \cite{Supratak2017}, however, we experimented with a sequence length of \{10, 20, 30\} epochs to have a comprehensive comparison with the proposed SeqSleepNet. 
	
	\subsection{Experimental Results}
	
	\setlength\tabcolsep{2.25pt}
	\begin{table*}[!t]
		\caption{Performance obtained by the proposed SeqSleepNet, the developed baselines, and existing works on the MASS dataset. We mark the proposed SeqSleepNet in bold, the developed baselines in italic, and existing works in normal font.  SeqSleepNet-$L$ indicates a SeqSleepNet with sequence length of $L$, a similar notation is used for E2E-DeepSleepNet baseline.}
		\vspace{-0.2cm}
		\scriptsize
		\begin{center}
			\begin{tabular}{|>{\arraybackslash}m{0.1in}|>{\arraybackslash}m{1.05in}|>{\centering\arraybackslash}m{0.6in}|>{\centering\arraybackslash}m{0.5in}|>{\centering\arraybackslash}m{0.35in}|>{\centering\arraybackslash}m{0.2in}|>{\centering\arraybackslash}m{0.275in}|>{\centering\arraybackslash}m{0.2in}|>{\centering\arraybackslash}m{0.2in}|>{\centering\arraybackslash}m{0.2in}|>{\centering\arraybackslash}m{0.2in}|>{\centering\arraybackslash}m{0.25in}|>{\centering\arraybackslash}m{0.2in}|>{\centering\arraybackslash}m{0.2in}|>{\centering\arraybackslash}m{0.2in}|>{\centering\arraybackslash}m{0.2in}|>{\centering\arraybackslash}m{0.2in}|>{\centering\arraybackslash}m{0.2in}|>{\centering\arraybackslash}m{0.2in}|>{\centering\arraybackslash}m{0.25in}|>{\centering\arraybackslash}m{0in} @{}m{0pt}@{}}
				\cline{3-20}
				\multicolumn{2}{c|}{} & \multirow{2}{*}{\makecell{Method}} & \multirow{2}{*}{\makecell{Feature \\ type}} & \multirow{2}{*}{\makecell{Num. of\\subjects}} &  \multicolumn{5}{c|}{Overall metrics} & \multicolumn{5}{c|}{Class-wise sensitivity} & \multicolumn{5}{c|}{Class-wise selectivity} & \parbox{0pt}{\rule{0pt}{2ex+\baselineskip}} \\ [0ex]  	
				\cline{6-20}
				\multicolumn{2}{c|}{} & & &  & Acc. & $\kappa$ & MF1 & Sens. & Spec. & W & N1 & N2 & N3 & REM & W & N1 & N2 & N3 & REM & \parbox{0pt}{\rule{0pt}{2ex+\baselineskip}} \\ [0ex]  	
				\cline{1-20}
				
				\multirow{10}{*}{\begin{sideways}{\bf Multi-output Systems}\end{sideways}} & \emph{\bf SeqSleepNet-30} & ARNN + RNN & learned & 200 &  $\bm{87.1}$ & $\bm{0.815}$ & $\bm{83.3}$ & $\bm{82.7}$ & $\bm{96.2}$ & $89.0$ & $\bm{59.7}$ & $\bm{90.9}$ & $80.2$ & $\bm{93.5}$ & $\bm{90.7}$ & $\bm{65.1}$ & $88.9$ & $\bm{84.2}$ & $90.7$ & \parbox{0pt}{\rule{0pt}{0.5ex+\baselineskip}} \\ [0ex]  	
				
				& \emph{\bf SeqSleepNet-20} & ARNN + RNN & learned & 200  &  $\bm{87.0}$ & $\bm{0.815}$ & $\bm{83.3}$ & $\bm{82.8}$ & $\bm{96.3}$ & $\bm{89.4}$ & $\bm{60.8}$ & $\bm{90.7}$ & $80.3$ & $\bm{92.9}$ & $\bm{90.0}$ & $\bm{65.1}$ & $\bm{89.1}$ & $\bm{84.0}$ & $90.8$ & \parbox{0pt}{\rule{0pt}{0.5ex+\baselineskip}} \\ [0ex]  	
				
				& \emph{\bf SeqSleepNet-10} & ARNN + RNN & learned  & 200  &  $\bm{87.0}$ & $\bm{0.814}$ & $\bm{83.2}$ & $\bm{82.4}$ & $\bm{96.2}$ & $88.6$ & $\bm{59.9}$ & $\bm{91.2}$ & $79.4$ & $\bm{93.0}$ & $\bm{91.3}$ & $\bm{64.9}$ & $88.6$ & $\bm{85.1}$ & $90.2$ & \parbox{0pt}{\rule{0pt}{0.5ex+\baselineskip}} \\ [0ex]  	
				
				& {\it E2E-DeepSleepNet-30} & CNN + RNN & learned & 200  & $86.4$  & $0.805$ & $82.2$ & $81.8$ & $96.1$ & $89.2$ & $55.8$ & $90.5$ & $83.1$ & $90.3$ & $88.8$ & $62.6$ & $88.8$ & $82.0$ & $91.1$ & \parbox{0pt}{\rule{0pt}{0.5ex+\baselineskip}} \\ [0ex]  	
				
				& {\it E2E-DeepSleepNet-20} & CNN + RNN & learned & 200  & $86.2$  & $0.804$ & $82.2$ & $82.0$ & $96.1$ & $88.4 $ & $ 57.0 $ & $ 89.9 $ & $ \bm{84.1} $ & $ 90.4 $ & $ 89.0 $ & $ 62.1 $ & $ 89.0 $ & $ 81.1 $ & $ \bm{91.2} $ & \parbox{0pt}{\rule{0pt}{0.5ex+\baselineskip}} \\ [0ex]  	
				
				& {\it E2E-DeepSleepNet-10} & CNN + RNN & learned & 200  & $86.3$  & $0.804$ & $82.0$ & $81.6$ & $96.1$ & $ 88.4 $ & $ 55.6 $ & $ 90.3 $ & $ 83.4 $ & $ 90.6 $ & $ 88.8 $ & $ 62.0 $ & $ 89.0 $ & $ 82.3 $ & $ 90.2 $& \parbox{0pt}{\rule{0pt}{0.5ex+\baselineskip}} \\ [0ex]  	
				
				& {\it M-E2E-ARNN} & ARNN & learned & 200  & $83.8$  & $0.767$ & $77.7$ & $77.0$ & $95.3$ & $85.0  $ & $ 37.4 $ & $ 89.2 $ & $ 79.2 $ & $ 94.2 $ &  $ 86.5 $ & $ 61.4 $ & $ 86.5 $ & $ 82.6 $ & $ 81.9 $ & \parbox{0pt}{\rule{0pt}{0.5ex+\baselineskip}} \\ [0ex]  	
				
				& Multitask 1-max CNN \cite{Phan2018e}  & CNN & learned & 200  & $83.6$ & $0.766$ & $77.9$ & $77.4$ & $95.3$ & $84.6$ & $41.1$ & $88.5$ & $79.7$ & $93.3$ & $86.3$ & $55.2$ & $86.9$ & $83.0$ & $83.3$ & \parbox{0pt}{\rule{0pt}{0.5ex+\baselineskip}} \\ [0ex]  	
				
				& DeepSleepNet2 \cite{Supratak2017} & CNN + RNN & learned & 62 (SS3) & $86.2$ & $0.800$ & $81.7$ & - & - & - & - & - & - & - & - & - & - & - & - & \parbox{0pt}{\rule{0pt}{0.5ex+\baselineskip}} \\ [0ex]  	
				
				& Dong \emph{et al.} \cite{Dong2017}  & DNN + RNN & learned& 62 (SS3) & $85.9$ & - & $80.5$ & - & - & - & - & - & - & - & - & - & - & - & - & \parbox{0pt}{\rule{0pt}{0.5ex+\baselineskip}} \\ [0ex]  	
				
				\cline{1-20}
				
				\multirow{10}{*}{\begin{sideways}{\bf Single-output Systems}\end{sideways}} & {\it E2E-ARNN} & ARNN & learned & 200  & $83.6$  & $0.766$ & $78.4$ & $78.0$ & $95.3$ & $86.6  $ & $ 43.7 $ & $ 87.8 $ & $ 80.9 $ & $ 91.2 $ & $ 86.3 $ & $ 57.6 $ & $ 87.2 $ & $ 82.3 $ & $ 82.4 $ & \parbox{0pt}{\rule{0pt}{0.5ex+\baselineskip}} \\ [0ex]  	
				
				& 1-max CNN \cite{Phan2018e} & CNN & learned &  200  & $82.7$  & $0.754$ & $77.6$ & $77.8$ & $95.1$ & $84.8$ & $46.8$ & $86.4$ & $82.0$ & $88.6$ & $86.2$ & $49.8$ & $87.4$ & $80.2$ & $84.2$ \parbox{0pt}{\rule{0pt}{0.5ex+\baselineskip}} \\ [0ex]  	
				
				& Chambon \emph{et al.} \cite{Chambon2018}  & CNN  & learned & 200 & $79.9$ & $0.726$ & $76.7$ & $80.0$ & $95.0$ & $81.1$ & $64.2$ & $76.2$ & $89.6$ & $89.0$ & $86.7$ & $41.0$ & $92.4$ & $73.1$ & $82.6$ & \parbox{0pt}{\rule{0pt}{0.5ex+\baselineskip}} \\ [0ex]  	
				
				& DeepSleepNet1 \cite{Supratak2017} & CNN (only) & learned & 200  & $80.7$ & $0.725$ & $75.8$ & $75.5$ & $94.5$ & $80.0$ & $51.9$ & $85.5$ & $69.0$ & $91.1$ & $87.5$ & $46.2$ & $85.3$ & $84.9$ & $79.7$ \parbox{0pt}{\rule{0pt}{0.5ex+\baselineskip}} \\ [0ex]  	
				
				& Tsinalis \emph{et al.} \cite{Tsinalis2016} & CNN & learned &  200  & $77.9$ & $0.680$ & $70.4$ & $69.4$ & $93.5$ & $82.3$ & $30.5$ & $86.8$ & $61.7$ & $85.8$ & $77.5$ & $44.7$ & $80.6$ & $80.0$ & $80.0$ & \parbox{0pt}{\rule{0pt}{0.5ex+\baselineskip}} \\ [0ex]  	
				
				& Chambon \emph{et al.} \cite{Chambon2018}  & CNN & learned & 61 (SS3) & $83.0$ & - & - & - & - &  - & - & - & - & - & - & - & - & - & - & \parbox{0pt}{\rule{0pt}{0.5ex+\baselineskip}} \\ [0ex]  	
				
				& DeepSleepNet1 \cite{Supratak2017} & CNN (only) & learned & 62 (SS3)  & $81.5$ & - & - & - & - & - & - & - & - & - & - & - & - & - & - & \parbox{0pt}{\rule{0pt}{0.5ex+\baselineskip}} \\ [0ex]  	
				
				& Dong \emph{et al.} \cite{Dong2017} & DNN (only) & learned & 62 (SS3)  & $81.4$ & - & $77.2$ & - & - & - & - & - & - & - & - & - & - & - & - & \parbox{0pt}{\rule{0pt}{0.5ex+\baselineskip}} \\ [0ex]  	
				
				& Dong \emph{et al.} \cite{Dong2017} & RF & hand-crafted & 62 (SS3)  & $81.7$ & - & $72.4$ & - & - & - & - & - & - & - & - & - & - & - & - & \parbox{0pt}{\rule{0pt}{0.5ex+\baselineskip}} \\ [0ex]  	
				
				& Dong \emph{et al.} \cite{Dong2017} & SVM & hand-crafted & 62 (SS3)  & $79.7$ & - & $75.0$ & - & - & - & - & - & - & - & - & - & - & - & - & \parbox{0pt}{\rule{0pt}{0.5ex+\baselineskip}} \\ [0ex]  	
				\cline{1-20}
			\end{tabular}
		\end{center}
		\label{tab:performance}
	\end{table*}

	\subsubsection{Sleep stage classification performance}
	
	We show in Table \ref{tab:performance} a comprehensive performance comparison of the proposed SeqSleepNet, the developed baselines, as well as published results on the MASS dataset. We report performance of a system using overall metrics, including accuracy, macro F1-score (MF1), Cohen's kappa ($\kappa$), sensitivity, and specificity. Performance on individual sleep stages are also assessed via class-wise sensitivity and selectivity as recommended in \cite{Imtiaz2014}. The systems are grouped into single-output or multiple-output to ease the interpretation.
	
	{\bf Impact of short-term sequential modelling.}
	The efficiency of short-term sequential modelling is highlighted by the superior performance of the E2E-ARNN baseline over those of the single-output systems. Compared to the best single-output CNN opponent (i.e. 1-max CNN \cite{Phan2018e}) on the entire MASS dataset, the E2E-ARNN baseline yields improvements of $0.9\%$ on overall accuracy. It also largely outperforms other single-output CNN architectures by $2.9\%$ to $5.7\%$. Performance gains can also be consistently seen on other metrics. It should be highlighted that the E2E-ARNN baseline adheres to the very standard one-to-one classification setup and does not make use of contextual input with multiple epochs as in many other CNN opponents, such as those proposed by Chambon \emph{et al.} \cite{Chambon2018} and Tsinalis \emph{et al.} \cite{Tsinalis2016}.
	
	\begin{figure*} [!t]
			\centering
			\includegraphics[width=1\linewidth]{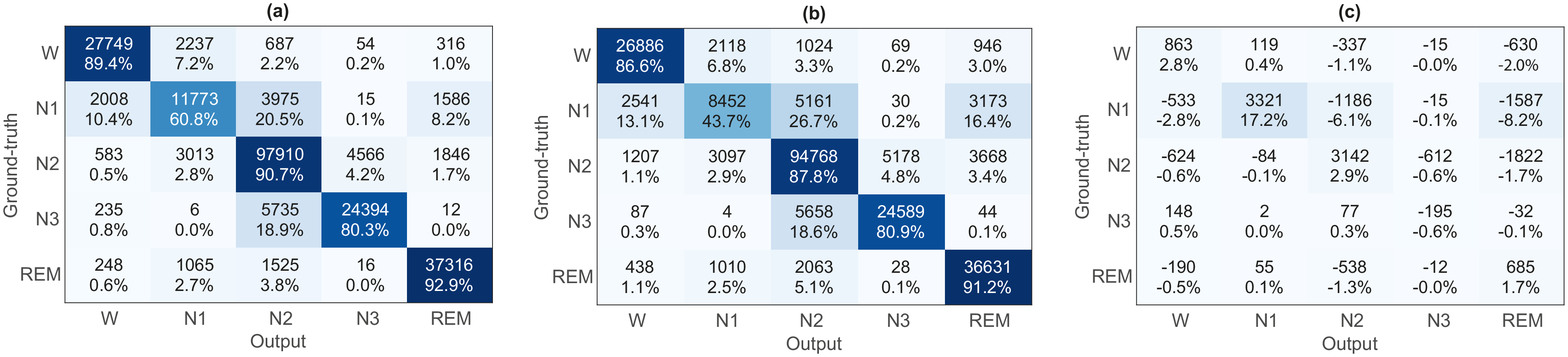}
			\caption{(a) The confusion matrix of SeqSleepNet-20 ($C_1$), (b) the confusion matrix of the E2E-ARNN baseline ($C_2$), and (c) the difference of two confusion matrices $C_1 - C_2$.}
			\label{fig:confusion_matrix_difference}
	\end{figure*}
	
	{\bf Single output vs multiple output.}
	Comparing the multi-output systems, the proposed SeqSleepNet outperforms other systems and set state-of-the-art performance on the MASS dataset with an overall accuracy, MF1, and $\kappa$ of $87.1\%$, $83.3\%$, and $0.815$, respectively. On the entire MASS dataset, it leads to an accuracy gain of $0.7\%$ absolute over the E2E-DeepSleepNet baseline which is the best competitor. Given that the top recurrent layers behave similarly on two networks (although SeqSleepNet has only one recurrent layer on the sequence level as well as smaller size of hidden state vectors), the improvement is likely due to the good epoch-wise sequential features learned by the epoch-level processing block of SeqSleepNet. On individual sleep stages, SeqSleepNet and the E2E-DeepSleepNet are comparable for Wake and N2 while the former shows its prominence on N1 which is usually very challenging to be recognized due to its similar characteristics to other stages and its low prevalence. Interestingly, in REM, SeqSleepNet is superior on sensitivity but inferior on selectivity compared to E2E-DeepSleepNet. This result suggests that SeqSleepNet is less conservative than E2E-DeepSleepNet on recognizing REM, i.e. it recognizes more but slightly lower-fidelity REM epochs. The opposite is observed on N3. Regarding the family of multitask networks, although the advantage of contextual output \cite{Phan2018e} is reflected by the improvement of these networks, i.e. the multitask CNN and the M-E2E-ARNN baseline, over their single-output peers, the limit of the contextual output size \cite{Phan2018e} makes their performance incomparable to those of the SeqSleepNet and the E2E-DeepSleepNet both of which can accommodate a much longer context, thanks to the capability of their sequence-level recurrent layers.
	
	{\bf Benefits of long-term sequential modelling.}
	The performance boost made by the proposed SeqSleepNet and the E2E-DeepSleepNet over their single-output counterparts also shed light into the power of long-term sequential modelling for automatic sleep staging. Averaged over all experimented sequence lengths, an accuracy gain of $3.4\%$ absolute is obtained by SeqSleepNet over the E2E-ARNN baseline. Likewise, an average accuracy improvement of $5.6\%$ yielded by the E2E-DeepSleepNet baseline over its bare CNN version (i.e. DeepSleepNet1 \cite{Phan2018e}) can also be seen. Previous works, e.g. Supratak \emph{et al.} \cite{Supratak2017} and Dong \emph{et al.} \cite{Dong2017} also presented a similar finding on the MASS subset SS3. However, the state-of-the-art performance of the proposed SeqSleepNet and the developed E2E-DeepSleepNet are obtained with end-to-end training, implying the unnecessity of multi-stage training \cite{Supratak2017,Dong2017}.
	
	In order to reveal the cause of improvement made by long-term sequential modelling, we further examine its effects on performances of individual classes. To this end, we computed the confusion matrix of the proposed SeqSleepNet with the sequence length of $L=20$ (denoted as $C_1$), the confusion matrix of the E2E-ARNN baseline (denoted as $C_2$), and inspect the difference between them, i.e. $C_1 - C_2$. In $C_1-C_2$, both positive diagonal entries and negative off-diagonal entries indicate improvements of SeqSleepNet over the E2E-ARNN baseline. It turns out that, long-term sequential modelling results in significant improvement on N1 with its accuracy boosted by $17.2\%$ while subtle influence is seen on other sleep stages. This achieved accuracy on the challenging N1 stage is also better than those reported in previous works \cite{Phan2018e,Supratak2017,Dong2017,Chambon2018}. These results suggests that long-term sequential modelling is more important than specific changes in the sleep stages.
	
	\subsubsection{Hypnogram}
\begin{figure*} [!t]
		\centering
		\includegraphics[width=1\linewidth]{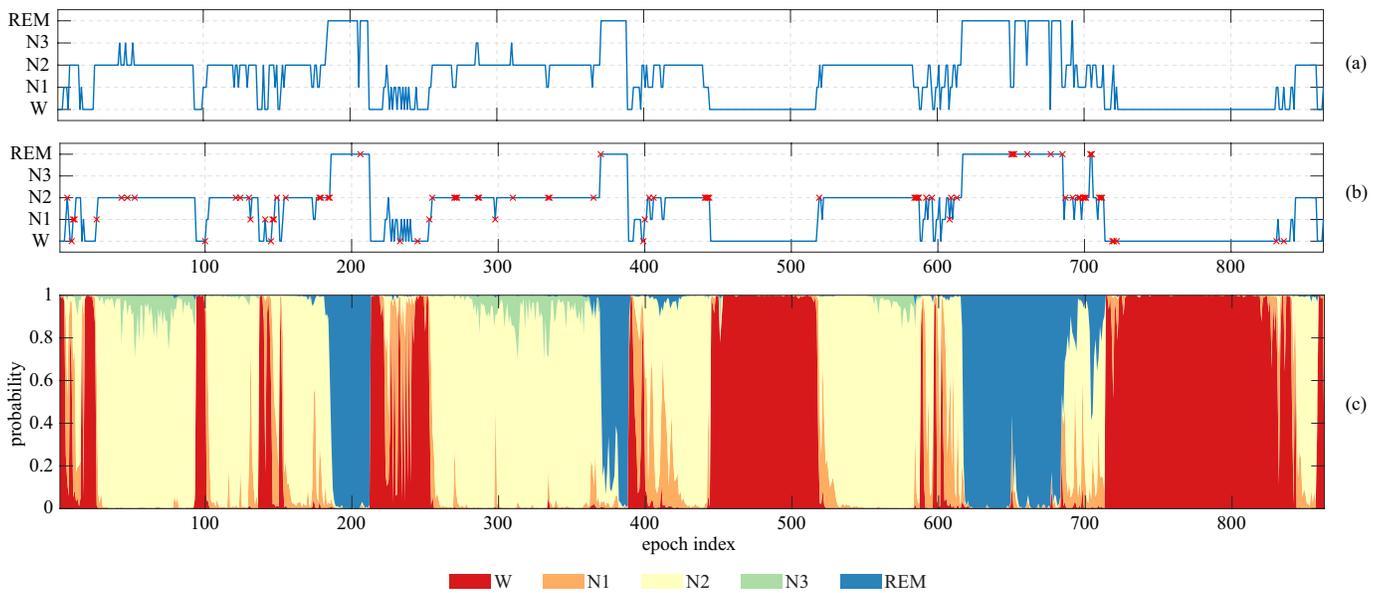}
		\caption{Output hypnogram (a) produced by the proposed SeqSleepNet ($L=20$) for subject 22 of the MASS dataset compared to the ground-truth (b). The errors are marked by the \textcolor{red}{$\times$} symbol. The posterior probability distribution over different sleep stages is shown in (c).}
		\label{fig:hypnogram}
	\end{figure*}

	Figure \ref{fig:hypnogram} further shows the output hypnogram and the posterior probability distribution per stage of sleep of a subject of the MASS dataset (subject 22 of subset SS1). It can be seen that the output hypnogram aligns very well with the corresponding ground truth. Often, the network makes errors at the short stage transition epochs. More specifically, on the entire MASS dataset, out of misclassified epochs made by SeqSleepNet-20, $44.0\%$ are transitioning and the rest $56.0\%$ are non-transitioning. However, when we inspected the transitioning set (constituting $16.6\%$ of the data) and the non-transitioning set (constituting $83.4\%$ of the data) seperately, an error rate of $34.5\%$ is seen on the former whereas that of the latter is four times lower, only $8.7\%$. This result suggests that the transitioning epochs are much harder to correctly classified compared to the non-transitioning ones. The rationale is that the transitioning epochs often contain information of two or three sleep stages, not to mention that the way we converted 20-second epochs to 30-second ones (cf. Section \ref{sec:datasets}) makes the stage overlap even worse. As a result, these present stages are active as indicated in the probability distribution in Figure \ref{fig:hypnogram}, however, we had to pick one of them as the final discrete output label for the sleep staging task.

\begin{table}[b!]
	\centering
	\caption{Influence of SeqSleepNet's recurrent depth on the overall accuracy.}
	\begin{tabular}{|>{\centering\arraybackslash}m{0.75in}|>{\centering\arraybackslash}m{0.5in}|>{\centering\arraybackslash}m{0.5in}|>{\centering\arraybackslash}m{0.5in}|>{\centering\arraybackslash}m{0in} @{}m{0pt}@{}}
		\cline{1-4}
		\multirow{2}{*}{\makecell{Recurrent depth}} & \multicolumn{3}{c|}{Sequence length} & \parbox{1pt}{\rule{0pt}{1ex+\baselineskip}} \\ [0ex]  	
		\cline{2-4}
		&  $L=10$ &  $L=20$ &  $L=30$ & \parbox{1pt}{\rule{0pt}{1ex+\baselineskip}} \\ [0ex]  	
		\cline{1-4}
		1 & $87.0$ & $87.0$ & $87.1$ & \parbox{1pt}{\rule{0pt}{0.5ex+\baselineskip}} \\ [0ex]
		2 & $86.8$ & $87.0$ & $87.1$ & \parbox{1pt}{\rule{0pt}{0.5ex+\baselineskip}} \\ [0ex]
		\cline{1-4}
	\end{tabular}
	\label{tab:influence_deepness}
	\vspace{0.1cm}
\end{table}
	
	\subsubsection{Influence of the sequence length and the network's depth}
	
	It can be seen from the results in Table \ref{tab:performance} that the sequence length equal or greater than 10 has minimal impact on the network performance. This observation is generalized for both SeqSleepNet and the E2E-DeepSleepNet as their accuracies vary in a negligible margin of $0.1\%$ when $L=\{10, 20, 30\}$.
	
	We carried out an additional experiment to study the influence of the deepness of SeqSleepNet's recurrent layers. We constructed the SeqSleepNet with two layers for both its epoch-level and sequence-level recurrent layers. A deep RNN was formed by stacking the GRU cells one on another as in \cite{phan2017c,phan2018d}. The overall accuracy of this network is shown in Table \ref{tab:influence_deepness} alongside that of the SeqSleepNet which has recurrent depth of 1. The results reveal that increasing the number of recurrent layers does not change the network's accuracy when the sequence length is sufficiently large, i.e. $L=20,30$. With $L=10$, an accuracy drop of $0.2\%$ is noticeable. A possible explanation is that, with short sequence length, the stronger network with the recurrent depth of 2 is more prone to overfitting than the simpler one with the recurrent depth of 1. This effect is not observed with larger sequence lengths as heavier multitasking helps to regularize the networks better.

	\subsubsection{Visualization of the learned attention weights}
\begin{figure*} [!t]
		\centering
		\includegraphics[width=1\linewidth]{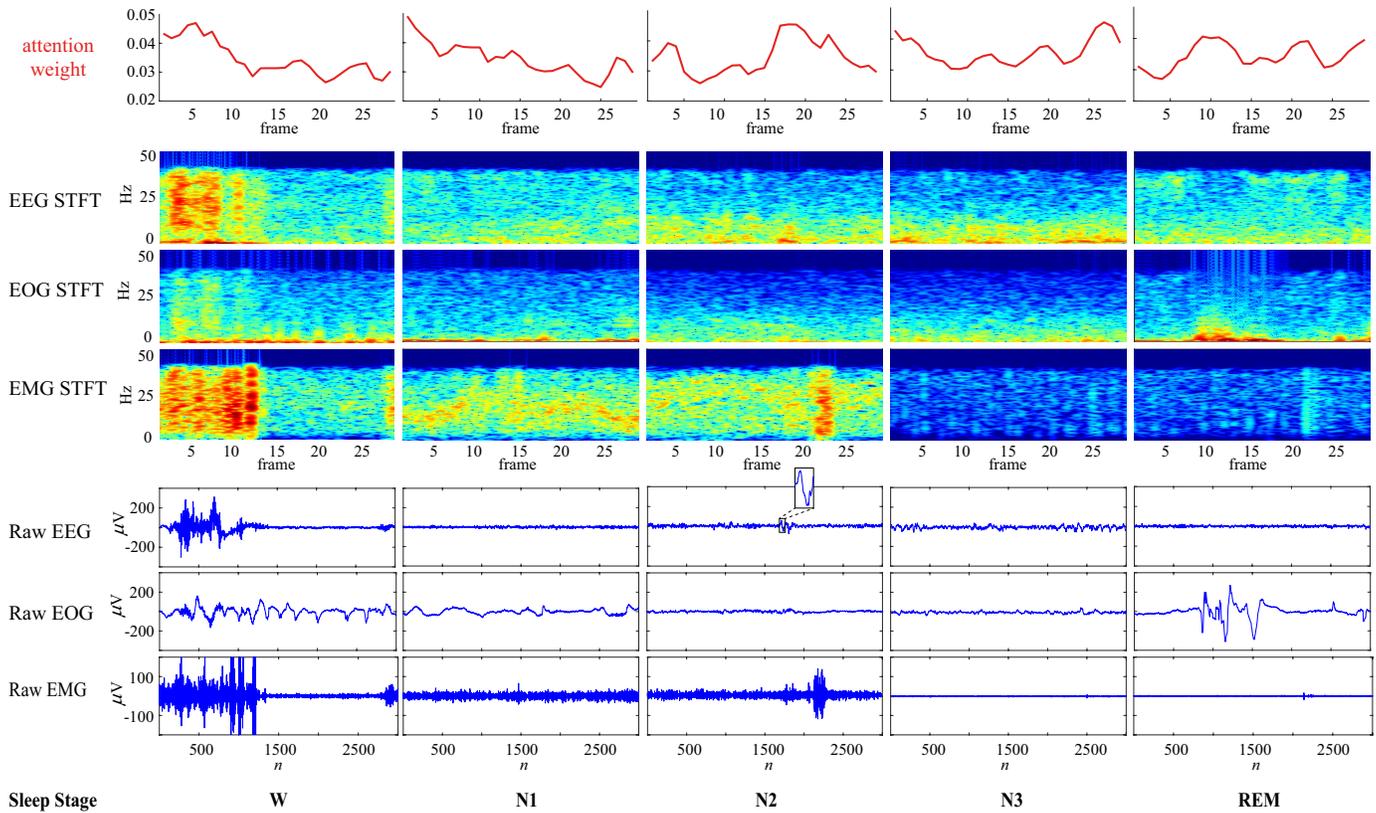}
		\caption{Attention weight learned by SeqSleepNet ($L=20$) for specific epochs of different sleep stages. Note that we generated the spectrograms with finer temporal resolution (2-second window with 90\% overlap) for visualization purpose.}
		\label{fig:attention_weights}
		\vspace{-0.1cm}
	\end{figure*}

	To shed light on how the SeqSleepNet has picked up features to distinguish one sleep stage from others, Figure \ref{fig:attention_weights} shows the attention weights for five specific epochs of different sleep stages. As expected, for the Wake epoch, the attention weights are particularly large in the region of high brain activities and muscle tone which are common characteristics discriminating Wake against other sleep stages. Similarly, for the REM epoch, more attention weights are put on ocular activities which are REM representative. Interestingly, attention layers also capture typical features of the N2 and N3 epoch as stronger weights are seen with occurrences of K-complex and slow brain waves, respectively.
	
	\section{Discussion}
	\label{sec:discussion}
	
	With the good performance demonstrated, the proposed SeqSleepNet has the potential to automate and replace manual sleep scoring \cite{Iber2007, Hobson1969}. Although SeqSleepNet's overall performance is just approximately $1\%$ better than that of the runner-up DeepSleepNet, it is worth noticing that this improvement is not evenly distributed over all sleep stages (cf. Table \ref{tab:performance}). While the networks perform more or less comparably on some stages (e.g. N2 and Wake), SeqSleepNet significantly outperforms DeepSleepNet on other stages (e.g. N1 and REM). This result might also be clinically meaningful as performing well on N1 and REM sleep makes SeqSleepNet potentially useful for diagnosis and assessments of many types of sleep disorders, such as narcolepsy \cite{AASM2014} and REM-Sleep Behavior Disorder (RBD) \cite{Cooray2018}. It is unlikely that SeqSleepNet trained on the MASS dataset, a cohort of healthy subjects, would directly work well on subjects with sleep disorders due to their different sleep architectures and characteristics compared to the healthy controls. However, a SeqSleepNet pre-trained with a large healthy cohort like the MASS dataset could serve as a starting point to be finetuned for another cohort of sleep pathologies, especially when the target cohort is of small size.
	
	SeqSleepNet also comes with some disadvantages. First, as a sequence-to-sequence model, the network needs to access entire sequences of multiple epochs to perform classification. This could delay online and realtime applications, such as sleep monitoring \cite{Mikkelsen2018b,Looney2016}. Second, the class-wise results in Table II show opposite behaviors of SeqSleepNet and DeepSleepNet on N3 and REM. This suggests that DeepSleepNet could compensate SeqSleepNet to improve performance on these two stages. It is therefore worth exploring their possible combinations to leverage their respective advantages.
	
	\section{Conclusions}
	\label{sec:conclusion}
	
	We proposed to treat automatic sleep staging as a sequence-to-sequence classification problem to jointly classify a sequence of multiple epochs at once. We then introduced a hierarchical recurrent neural network, i.e. SeqSleepNet, running on multichannel time-frequency image input to tackle this problem. The network is composed of parallel filterbank layers for preprocessing the image input, an epoch-level attention-based bidirectional RNN layer to encode sequential information of individual epochs, and a sequence-level bidirectional RNN layer to model inter-epoch sequential information. The network was trained end-to-end via dynamic folding and unfolding the input sequence at different levels of network hierarchy. We show that while sequential features learned for individual epochs by the epoch-level attention-based bidirectional RNN are more favourable than those learned by different CNN opponents, further capturing the long-term dependency between epochs by the top RNN layer leads to significant performance improvement. The proposed SeqSleepNet outperforms not only existing works but also the strong baselines developed for comparison, setting state-of-the-art performance on the entire MASS dataset.
	
	\section*{Acknowledgement}
	The research was supported by the NIHR Oxford Biomedical Research Centre, Wellcome Trust (grant  098461/Z/12/Z), and the Engineering and Physical Sciences Research Council (EPSRC -- grant EP/N024966/1).
	
	\ifCLASSOPTIONcaptionsoff
	\newpage
	\fi

	
	\balance
	
	%
	\bibliographystyle{IEEEtran}
	\bibliography{bibliography}

\end{document}